\documentclass[twocolumn]{IEEEtran}
\usepackage{graphicx,amssymb}
\begin{document}
\title{Blind Detection and Compensation of Camera Lens Geometric Distortions}
\author{Lili Ma, {\it Student Member, IEEE}, YangQuan Chen and Kevin L. Moore, {\it Senior Members, IEEE}\\
Center for Self-Organizing and Intelligent Systems (CSOIS),\\Dept. of Electrical and Computer Engineering, 4160 Old Main Hill,\\ Utah State University (USU), Logan, UT 84322-4160, USA.\\Email: %\texttt{moorek@ece.usu.edu}
\texttt{lilima@cc.usu.edu, \{yqchen, moorek\}@ece.usu.edu}} \maketitle{}

\begin{abstract}
This paper presents a blind detection and compensation technique for camera lens geometric distortions. The lens
distortion introduces higher-order correlations in the frequency domain and in turn it can be detected using
higher-order spectral analysis tools without assuming any specific calibration target. The existing blind lens
distortion removal method only considered a single-coefficient radial distortion model. In this paper, two coefficients
are considered to model approximately the geometric distortion.
%A two coefficient radial distortion model is also
%considered for fair comparison.
All the models considered have analytical closed-form inverse formulae.
\end{abstract}

%%%%%%%%%%%%%%%%%%%%%%%%%%%%%%%%%%%%%%%%%%%%%%%%%%%%%%%%%%%%%%%%%%%%%%%%%%%%%%%%%%%%%%%%%%%%%%%%%%%%%%%%%%%%
\section{Introduction}

Generally, lens detection and compensation is modelled as one step in camera calibration, where the camera calibration
is to estimate a set of parameters describing the camera's imaging process. Using a camera's pinhole model, the
projection from the 3-D space to the image plane can be described by
\begin{equation} \label{eqn: projection matrix}
\lambda \left [\matrix{u\cr v\cr 1} \right ] = \underbrace{\left[\matrix{\alpha & \gamma & u_0 \cr 0 & \beta & v_0\cr 0
& 0 & 1}\right]}_{A_{\rm intr}} \left[\matrix{X\cr Y\cr Z}\right] = Z A_{\rm intr} \left[\matrix{ X/Z \cr Y/Z \cr
1}\right],
\end{equation}
where the matrix $A_{\rm intr}$ denotes the camera's intrinsic matrix. $(\alpha, \gamma, \beta, u_0, v_0)$ are the
camera's five intrinsic parameters, with $(\alpha, \beta)$ being two scalars in the two image axes, $(u_0, v_0)$ the
coordinates of the principal point (which is also assumed to be at the center of distortion in this paper), and
$\gamma$ describing the skewness of the two image axes.
%Throughout this paper, we call $(u,v)$ the ideal points on
%the image plane, and $(x,y)= (X/Z, y = Y/Z)$ the normalized ideal points in the camera-centered frame (camera frame for
%short).
In the imaging process described by (\ref{eqn: projection matrix}), the effect of lens distortion has not been
considered.

In equation (\ref{eqn: projection matrix}), $(u,v)$ is not the actually observed image point since virtually all
imaging devices introduce certain amount of nonlinear distortions. Among the nonlinear distortions, radial distortion,
which is performed along the radial direction from the center of distortion, is the most severe part
\cite{OlivierF01Straight,tsai87AVersatile}. In the category of polynomial radial distortion modeling, the commonly used
radial distortion function is governed by the following polynomial equation
\cite{Photogrammetry80,zhang99calibrationinpaper,Heikkil97fourstepcameracalibration,Janne96Calibration,Undistortionchapter,Juyang92distortionmodel}:
\begin{equation} \label{eqn: general polynomial}
r_d = r + \delta_r = r \, f(r, {\bf k}) = r \, (1 + k_1 r^2 + k_2 r^4 + k_3 r^6 + \cdots),
\end{equation}
where ${\bf k} = [k_1, k_2, k_3, \ldots]$ is a vector of distortion coefficients. $r$ and $r_d$ can be defined either
in the camera frame as
\cite{Photogrammetry80,zhang99calibrationinpaper,Heikkil97fourstepcameracalibration,Janne96Calibration}
\begin{equation} \label{eqn: r rd in xy}
r_{\rm xy} = \sqrt{x^2 + y^2}, \quad r_{\rm dxy} = \sqrt{x_d^2 + y_d^2},
\end{equation}
with
\begin{equation}
x = \frac{X}{Z}, \quad y = \frac{Y}{Z},
\end{equation}
or on the image plane as \cite{Undistortionchapter,Juyang92distortionmodel}
\begin{equation} \label{eqn: r rd in uv}
\begin{array}{c}
r_{\rm uv} = \sqrt{(u - u_0)^2 + (v - v_0)^2}, \\[5pt]
r_{\rm duv} = \sqrt{(u_d - u_0)^2 + (v_d - v_0)^2},
\end{array}
\end{equation}
where the subscript $_d$ denotes the distorted version of the corresponding ideal projections. The subscripts $_{\rm
uv}$ and $_{\rm xy}$ denote the definitions on the image plane and the camera frame, respectively. The distortion
models discussed in this paper are in the Undistorted-Distorted formulation \cite{Toru02Unified}, though similar idea
can be applied to the Distorted-Undistorted formulation.

Most of the existing camera calibration and lens distortion compensation techniques require either explicit calibration
target, whose 2-D or 3-D metric information is available \cite{zhang99calibrationinpaper}, or an environment rich of
straight lines \cite{OlivierF01Straight}. The above mentioned techniques are suitable for situations where the camera
is available. For situations where direct access to the imaging device is not available, such as when down loading
images from the web, blind lens removal technique has been exploited based on frequency domain criterion
\cite{Hany01blindremoval}. The fundamental analysis is based on the fact that lens distortion introduces higher-order
correlations in the frequency domain, where the correlations can be detected via tools from higher-order spectral
analysis (HOSA). However, it has been reported that the accuracy of blindly estimated lens distortion is by no means
comparable to those based on calibration targets. Due to this reason, this approach can  be useful in areas where only
qualitative results are required \cite{Hany01blindremoval}.

The paper is organized as follows. Section~\ref{sec: blind technique} starts with an introduction of the blind lens
distortion removal technique, which is based on the detection of higher-order correlations in the frequency domain. In
Sec.~\ref{sec: lens distortion modeling}, we briefly describe two existing polynomial/rational radial distortion
approximation functions, along with our recently developed simplified geometric distortion modeling method.
Section~\ref{sec: experiments} presents detailed lens distortion compensation results using images from both calibrated
cameras and some web images. The previously calibrated cameras act as a validation of the blind lens removal technique.
Finally, section~\ref{sec: conclusions} concludes the paper.

%%%%%%%%%%%%%%%%%%%%%%%%%%%%%%%%%%%%%%%%%%%%%%%%%%%%%%%%%%%%%%%%%%%%%%%%%%%%%%%%%%%%%%%%%%%%%%%%%%%%%%%%%%%%
\section{Frequency Domain Blind Lens Detection}
\label{sec: blind technique}

In this section, higher-order spectral analysis is first reviewed, which provides the fundamental criterion for the
blind lens distortion removal technique \cite{Hany01blindremoval,Mendel91HOS}. The basic approach of the blind lens
distortion removal exploits the fact that lens distortion introduces higher-order correlations in the frequency domain,
which can be detected using HOSA tools.

\subsection{Bispectral Analysis}

Higher-order correlations introduced by nonlinearities can be estimated by higher-order spectra \cite{Mendel91HOS}. For
example, third-order correlations can be estimated by bispectrum, which is defined as
\begin{equation} \label{eqn: bispectrum}
B(\omega_1, \omega_2) = E \{F(\omega_1) F(\omega_2) F^*(\omega_1 + \omega_2)\},
\end{equation}
where $E\{\cdot\}$ is the expected value operator and $F(\omega)$ is the Fourier transform of a stochastic
one-dimensional signal in the form of
\begin{equation} \label{eqn: power spectrum}
F(\omega) = \sum^{\infty}_{k = -\infty} f(k) e^{-j \omega k}.
\end{equation}
Notice that the bispectrum of a real signal is complex-valued. Since the estimate of the above bispectrum has the
undesired property that its variance at each bi-frequency $(\omega_1, \omega_2)$ is dependent of the bi-frequency, a
normalized bispectrum, called the bicoherence, is exploited, which is defined to be
\cite{Hany01blindremoval,Mendel91HOS}
\begin{equation}\label{eqn: bicoherence}
b^2(\omega_1, \omega_2) = \frac{|B^2(\omega_1, \omega_2)|}{E\{|F(\omega_1) F(\omega_2)|^2\} \,
E\{|F(\omega_1+\omega_2)|^2\}}.
\end{equation}
The above bicoherence can be estimated as
\begin{equation} \label{eqn: bicoherence estimator}
{\hat b}(\omega_1, \omega_2) = \frac{\frac{1}{N} \sum_k F_k(\omega_1) F_k (\omega_2) F^{*}_k
(\omega_1+\omega_2)}{\sqrt{\frac{1}{N} \sum_k |F_k(\omega_1) F_k(\omega_2)|^2 \frac{1}{N} \sum_k
|F_k(\omega_1+\omega_2)|^2}},
\end{equation}
which becomes a real-valued quantity. As a measure of the overall correlations, the following quantity is employed in
\cite{Hany01blindremoval}
\begin{equation} \label{eqn: sum bicoherence measure}
\frac{1}{N^2} \sum^{N/2}_{\omega_1 = -N/2} \sum^{N/2}_{\omega_2 = -N/2} {\hat b} \left( \frac{2 \pi \omega_1}{N},
\frac{2 \pi \omega_2}{N}\right),
\end{equation}
where $N$ is the dimension of the input one-dimensional signal.

%%%%%%%%%%%%%%%%%%%%%%%%%%%%%%%%%%%%%%%%%
\subsection{Blind Lens Removal Algorithm}
\label{sec: blind removal procedures}

Consider a signal $f_d(x)$ that is a distorted version of $f(x)$ according to
\begin{equation}
f_d(x) = f(x(1 + \kappa x^2)),
\end{equation}
with $\kappa$ controlling the amount of distortion. It has been shown in \cite{Hany01blindremoval} that correlations
introduced by the nonlinearity is proportional to the distortion coefficient $\kappa$, where the quantity (\ref{eqn:
sum bicoherence measure}) is chosen as the measure of the correlations. Now, consider the inverse problem of recovering
$f(x)$ from $f_d(x)$. It is only when $\kappa$ is properly estimated that the inverted ${\hat f}(x)$ contains a least
amount of nonlinearities, in which case ${\hat f}(x)$ holds a minimum bicoherence.

Based on the above discussions, an intuitive algorithm applied for the blind lens distortion removal is listed in the
following \cite{Hany01blindremoval}:
\begin{itemize}
\item [\rm \bf 1)] Select a range of possible values for the distortion coefficients $\bf \kappa$. \item [\rm \bf 2)]
For each $\bf \kappa$, perform inverse undistortion to $f_d(x)$ yielding a provisional undistortion function $f_{\bf
\kappa}(x)$. \item [\rm \bf 3)] Compute the bicoherence of $f_{\bf \kappa}(x)$. \item [\rm \bf 4)] Select the $\bf
\kappa$ that minimizes all the calculated bicoherence of the undistorted signals. \item [\rm \bf 5)] Remove the
distortion using the distortion coefficient obtained from step {\bf 4)}.
\end{itemize}

%%%%%%%%%%%%%%%%%%%%%%%%%%%%%%%%%%%%%%%%%%%%%%%%%%%%%%%%%%%%%%%%%%%%%%%%%%%%%%%%%%%%%%%%%%%%%%%%%%%%%%%%%%%%
\section{Lens Distortion Modeling}
\label{sec: lens distortion modeling}

The radial distortion model applied for the blind lens distortion removal in \cite{Hany01blindremoval} is
\begin{equation} \label{eqn: one coeff}
r_{\rm duv} =  r_{\rm uv} (1 + k_{\rm uv} \, r^2_{\rm uv}),
\end{equation}
which is equivalent to
\begin{equation}
\begin{array}{c}
(u_d - u_0) =  (u - u_0) (1 + k_{\rm uv} \, r^2_{\rm uv}), \\[5pt]
(v_d - v_0) =  (v - v_0) (1 + k_{\rm uv} \, r^2_{\rm uv}),
\end{array}
\end{equation}
when assuming the center of distortion is at the principal point.

%The radial distortion modeling $1 + k \, r^2_{\rm uv}$ is a special case of (\ref{eqn: general polynomial}) with one
%coefficient. A set of distortion functions have been investigated in
%\cite{LiliSub2ITMI03Rational,LiliSub2IT03geometric} for satisfactory calibration accuracy along with the desirable
%property of having analytical inverse formulae. As an example, with two coefficients, the following distortion function
%can be applied to model the radial distortion:
%\begin{equation} \label{eqn: quadratic function}
%f(r_{\rm uv}, {\bf k}) = 1 + k_{\rm uv1} r_{\rm uv} + k_{\rm uv2} r_{\rm uv}^2,
%\end{equation}
%which has been shown in \cite{LiliSub2ITMI03Rational} to have analytical inverse formulae with better calibration
%accuracy compared to (\ref{eqn: one coeff}).

Besides the radial distortion modeling, we have recently proposed a simplified geometric lens distortion modelling
method, where lens distortion on the image plane can be modelled by \cite{LiliSub2IT03geometric}
\begin{equation} \label{eqn: U-D model in udvd}
\begin{array}{c}
(u_d - u_0) = (u - u_0) \, f(r_{\rm uv}, {\bf k}_{\rm uv1}), \\[5pt]
(v_d - v_0) = (v - v_0) \, f(r_{\rm uv}, {\bf k}_{\rm uv2}),
\end{array}
\end{equation}
where $f(r_{\rm uv}, {\bf k}_{uv 1,2})$ can be chosen to be any of the available distortion functions. To illustrate
the simplified geometric distortion idea, in this paper, $f(r_{\rm uv}, {\bf k}_{uv1})$ and $f(r_{\rm uv}, {\bf
k}_{uv2})$ are chosen to be the following rational functions \cite{LiliSub2ITMI03Rational}:
\begin{equation} \label{eqn: simplified geometric 6}
\begin{array}{c}
\displaystyle f(r_{\rm uv}, k_{\rm uv1}) = \frac{1}{1 + k_{\rm uv1} r_{\rm uv}^2},\\
\displaystyle f(r_{\rm uv}, k_{\rm uv2}) = \frac{1}{1 + k_{\rm uv2} r_{\rm uv}^2},
\end{array}
\end{equation}
for its fewer number of distortion coefficients and the property of having analytical geometric undistortion formulae.
From equations (\ref{eqn: U-D model in udvd}) and (\ref{eqn: simplified geometric 6}), we have
\begin{equation} \label{eqn: geometric undistortion 6}
(u_d - u_0)^2 (1 + k_{\rm uv1} {\bar r}_{\rm uv})^2 + (v_d - v_0)^2 (1+k_{\rm uv2} {\bar r}_{\rm uv})^2 = {\bar r}_{\rm
uv},
\end{equation}
with ${\bar r}_{\rm uv} \stackrel{\Delta}{=} r^2_{\rm uv}$. The above equation is a quadratic function in ${\bar
r}_{\rm uv}$, thus having analytical inverse formula.

%%%%%%%%%%%%%%%%%%%%%%%%%%%%%%%%%%%%%%%%%%%%%%%%%%%%%%%%%%%%%%%%%%%%%%%%%%%%%%%%%%%%%%%%%%%%%%%%%%%%%%%%%%%%
\section{Experimental Results}
\label{sec: experiments}

% First, verify if blind removal technique really works:
In this section, we first verify that the blind lens removal technique, which is based on the detection of higher-order
correlations in the frequency domain, can be used for the detection and compensation for lens distortion. This
verification is via the comparison of the calibration coefficients of the blind removal technique with those calibrated
by a planar-target based calibration method \cite{zhang99calibrationinpaper}. Though, the calibration results by the
blind removal technique are by no means comparable to those based on calibration target, the results shown in
Sec.~\ref{sec: blind verification} does manifest a reasonable accuracy, at least for applications where only
qualitative performance is required.

% Be able to detect geometric distortion, at least showing the trend:
The existing blind lens distortion removal method only considered a single-coefficient radial distortion model, as
described in (\ref{eqn: one coeff}). In Sec.~\ref{sec: blind geometric}, we show that cameras, which are more
accurately modelled by different distortion coefficients along the two image axes, can also be detected using
higher-order correlations. As an example, the simplified geometric distortion modeling with the distortion function
(\ref{eqn: simplified geometric 6}) is applied. Using single-coefficient to describe the distortion along each image
axis, totally two coefficients are used in (\ref{eqn: simplified geometric 6}). One reason for choosing the function
(\ref{eqn: simplified geometric 6}) is in its fewer number of distortion coefficients that reduces the whole
optimization duration. Another advantage lies in that equation (\ref{eqn: simplified geometric 6}) has analytical
geometric inverse formula, which further advances the optimization speed.

% Lens compensation of web images:
Finally, lens distortion compensation of several web images are performed using the blind removal technique.

%===========================================================================
\subsection{Verification of Blind Lens Compensation Using Calibrated Cameras}
\label{sec: blind verification}
% two calibrated cameras: desktop, ODIS cameras:
% For radial distortion modeling: compare the distortion coefficient estimated by the blind technique with that obtained by calibration
% show the undistorted/compensated whole image for the distortion coefficients calibrated by the blind method and the calibration method

Comparisons between the distortion coefficients calibrated by the blind removal technique with those obtained by the
target-based calibration method have been performed in \cite{Hany01blindremoval} using distortion function (\ref{eqn:
one coeff}). Similar comparison is given here via two calibrated cameras, the desktop \cite{Lilicalreport02} and the
ODIS camera \cite{odiscamera}, using the distortion function (\ref{eqn: one coeff}), along with the function (\ref{eqn:
simplified geometric 6}) with $k_{\rm uv1} = k_{\rm uv2}$. We think that this double verification is needed since lens
nonlinearity detection using higher-order spectral analysis is a recent development.

\begin{figure}[htb]
\centering
\includegraphics[width=0.5\textwidth]{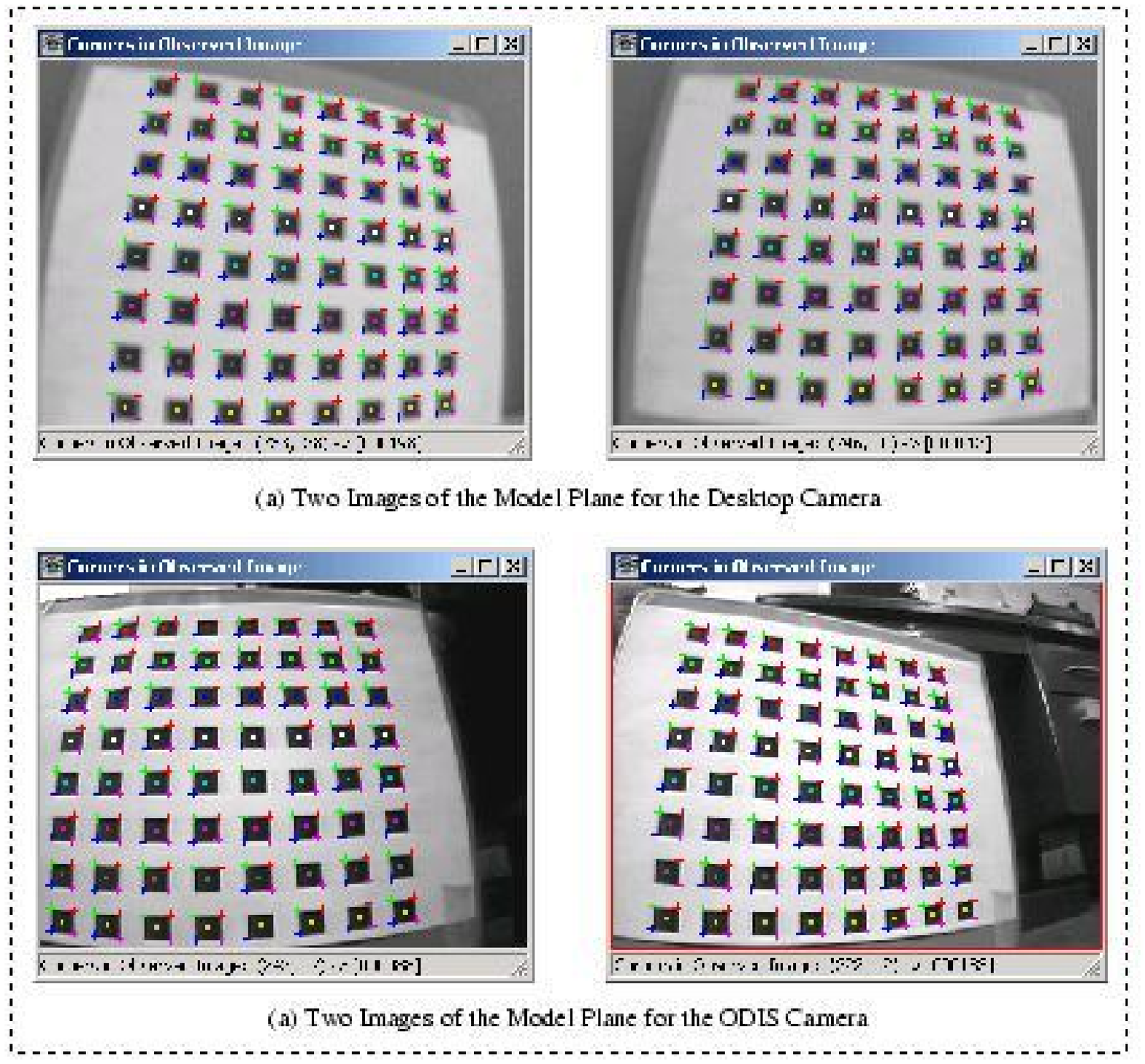}
\caption {Two Sample images of the model plane with the extracted corners (indicated by cross) for the desktop and ODIS
cameras.} \label{fig: extracted desktop ODIS}
\end{figure}

\begin{figure}[htb]
\centering
\includegraphics[width=0.5\textwidth]{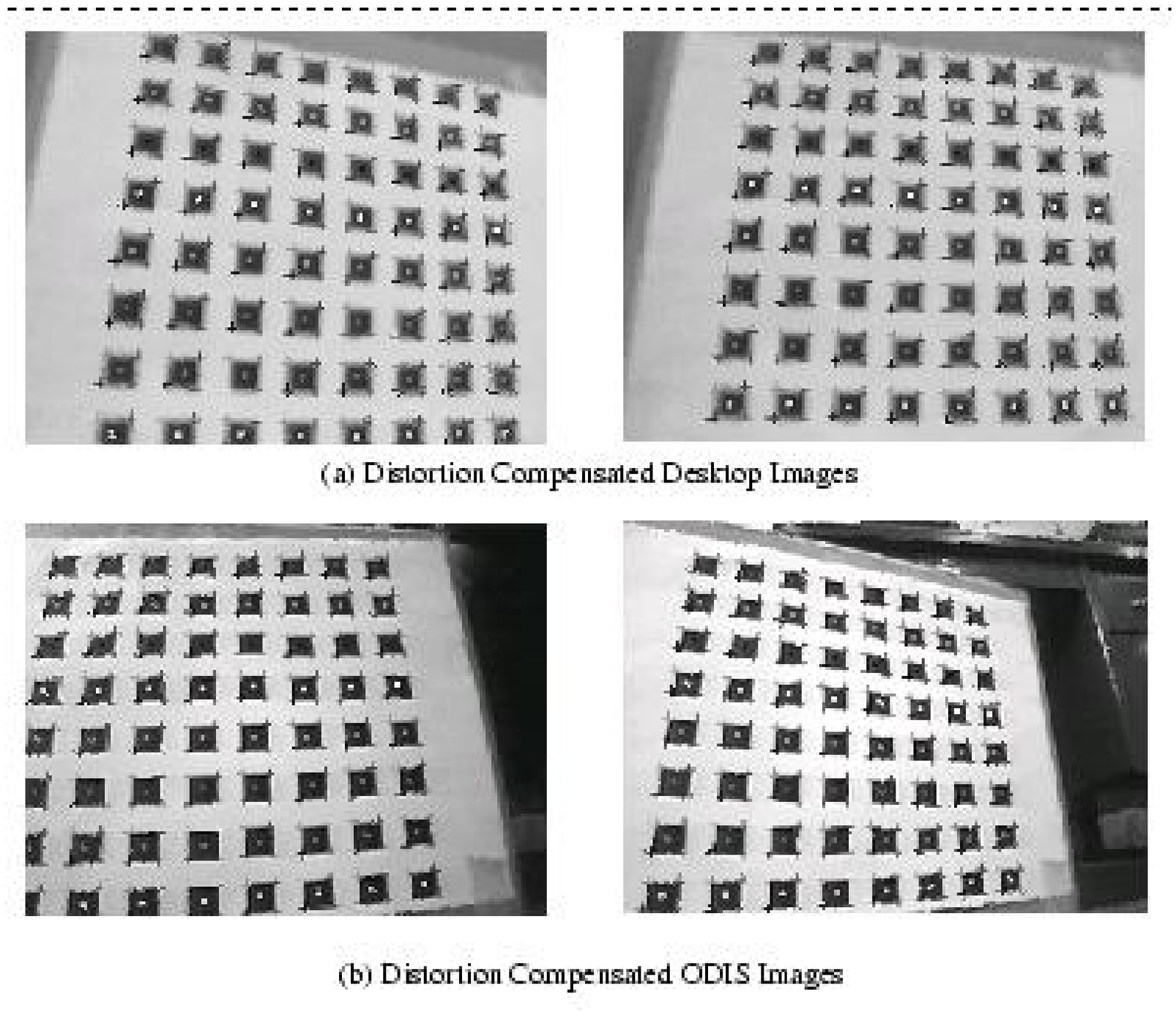}
\caption {Blindly radially compensated desktop and ODIS images with distortion function (\ref{eqn: simplified geometric
6}) with $k_{uv1} = k_{uv2}$.} \label{fig: blind compensated desktop ODIS 2}
\end{figure}

Original images of the desktop and the ODIS cameras are shown in Fig.~\ref{fig: extracted desktop ODIS}, where the
plotted dots in the center of each square are used for judging the correspondence with the world reference points for
the target-based calibration. Using the single-coefficient radial distortion model in (\ref{eqn: simplified geometric
6}) (with $k_{\rm uv1} = k_{\rm uv2}$), the blindly compensated images of the two cameras are shown in Fig.~\ref{fig:
blind compensated desktop ODIS 2}. An image interpolation operator is applied during the lens distortion compensation,
since the observed original images are shrinked due to the negative distortion coefficients. The undistorted images are
only plotted in gray-level to illustrate the lens compensation results\footnote{The image interpolation operator
currently applied is an average operator around each un-visited pixel in the compensated image. The resultant
undistorted images might have noise and blur due to this simple operator.}. Comparing Fig.~\ref{fig: extracted desktop
ODIS} with Fig.~\ref{fig: blind compensated desktop ODIS 2}, it can be observed that lens distortion is reduced
significantly, though not completely and perfectly.

Using the planar calibration target observed by the cameras as shown in Fig.~\ref{fig: extracted desktop ODIS}, the
desktop and ODIS cameras have been calibrated in \cite{LiliSub2IT03geometric,LiliSub2ITMI03Rational} using the
planar-based camera calibration technique described in \cite{zhang99calibrationinpaper}. However, the calibrated camera
parameters in \cite{LiliSub2IT03geometric,LiliSub2ITMI03Rational} are in the normalized camera frame, while the blind
removal technique deals with distortions in the image plane directly. A transformation between the lens distortion
coefficients in the camera frame and those in the image plane is thus needed.

A rough transformation is illustrated in the following using the obtained intrinsic parameters from the planar-target
based calibration technique. From equation (\ref{eqn: projection matrix}), we have
$$(u - u_0) = \alpha x + \gamma y, \quad (v - v_0)  = \beta y.$$
Assuming that
\begin{equation} \label{eqn: rough assumption}
\gamma \approx 0, \quad  \alpha \approx \beta,
\end{equation}
for a coarse approximation and using a single-coefficient radial distortion model (\ref{eqn: one coeff}), we have
\begin{equation}
r_{\rm duv} = \alpha \, r_{\rm dxy}, \quad r_{\rm uv} = \alpha \, r_{\rm xy}.
\end{equation}
The relationship between $k_{\rm uv}$ and $k_{\rm xy}$ can be determined straightforward as
\begin{equation}
k_{\rm uv} = k_{\rm xy}/\alpha^2.
\end{equation}

In this paper, blind lens distortion compensation is implemented via Matlab using higher-order spectral analysis
toolbox following the procedures listed in Sec.~\ref{sec: blind removal procedures}. However, instead of using equation
(\ref{eqn: sum bicoherence measure}) as the objective function to minimize, the maximum value of bicoherence is used in
our implementation as the measurement criterion for nonlinearity, which is\footnote{The reason to use the criterion in
(\ref{eqn: max bicoherence measure}) is more experimental. In our simulations, distortion coefficients obtained via the
average sum criterion in (\ref{eqn: sum bicoherence measure}) deviates from the previously calibrated coefficients
significantly. However, when using the maximum bicoherence criterion (\ref{eqn: max bicoherence measure}), close and
reasonable calibration results can be obtained for both the desktop and the ODIS cameras.}
\begin{equation} \label{eqn: max bicoherence measure}
J = \frac{1}{N_1 N_2} {\bf \max}_{\omega_1 \in [-\frac{N}{2}, \frac{N}{2}], \; \omega_2 \in [-\frac{N}{2},
\frac{N}{2}]} {\hat b}\left( \frac{2 \pi \omega_1}{N}, \frac{2 \pi \omega_2}{N}\right),
\end{equation}
for an input image of dimension $N_1 \times N_2$.

Comparison of the distortion coefficients obtained from the blind removal and the target-based
\cite{zhang99calibrationinpaper} calibration techniques is shown in Table.~\ref{table: verification model 1} using the
functions (\ref{eqn: one coeff}) and (\ref{eqn: simplified geometric 6}) with $k_{\rm uv1} = k_{\rm uv2}$. It can be
observed from Table~\ref{table: verification model 1} that, despite the deviation of the blindly calibrated results
from those based on calibration targets, there is a consistency about the trend qualitatively. Notice that the
distortion coefficients under the ``Target'' column in Table~\ref{table: verification model 1} are obtained via
approximations, where of course more precise transformations can be achieved without applying the assumptions in
(\ref{eqn: rough assumption}). However, since currently the blind removal method considers only the lens distortion (no
calibration of the center of distortion), precise comparison can not be achieved even with precise calculation from the
side of the target-based algorithm. Generally, the comparison can only be performed ``quantitatively'', though the
blind removal technique does provide another quantitative criterion for evaluating the calibration accuracy.

\begin{table}[htb]
\centering \caption{Comparison of Lens Distortion Coefficients of the Blind Removal Technique and The Target-Based
Algorithm} \label{table: verification model 1}
\renewcommand{\arraystretch}{1.3}
\vspace{-2mm} {\small {\begin {tabular}{|c|c|c|c|c|}\hline & &\multicolumn{2}{|c|}{{\bf Blind Technique} ($10^{-6}$)} &
{\bf Target} \\\cline{3-4}

\raisebox{1.5ex}[0pt]{\bf Eqn.}& \raisebox{1.5ex}[0pt]{\bf Camera}& \bf Values & \bf Mean & ($10^{-6}$)\\\hline

& Desktop& $-[3.5,4.5,3.5,4.5,1.5]$ &-3.5 &$-3.73$ \\\cline{2-5}

\raisebox{1.3ex}[0pt]{(\ref{eqn: one coeff})}&ODIS    &$-[3.5, 2.5, 2.5, 3.5, 3.5]$& -3.1 &$-4.13$ \\\hline

& Desktop& $[5,6,5,5,1]$ &4.4 &$4.27$ \\\cline{2-5}

\raisebox{1.3ex}[0pt]{(\ref{eqn: simplified geometric 6})$^*$} &ODIS &$[5,3,4,5,4]$ & 4.2&$4.75$ \\\hline
\end {tabular}}}
{\small
\begin{minipage}{0.5\textwidth}
\vspace{1mm} \hspace{2mm}$^*k_{uv1} = k_{uv2}$ for modeling the radial distortion.
\end{minipage}}
\end{table}

% Selection of searching range of the distortion coefficients: image is not normalized
One issue in the implementation is how to select the searching range for an image. While image normalization method is
commonly applied, in our simulation, the searching range is determined based on the image's dimension and the observed
judgement of radial or pincushion distortions in a non-normalized way. This is the reason why in Table~\ref{table:
verification model 1}, all the distortion coefficients are very small values. More specifically, consider the radial
distortion function (\ref{eqn: one coeff}) with the maximum possible distortion at the image boundary. Let $r = r_{\rm
max}$ and $r_d = \rho \, r_{\rm max}$, where $r_{\rm max}$ is defined to be $r_{\rm max} = \sqrt{u_0^2 + v_0^2}$ on the
image plane and the subscript $_{\rm uv}$ is dropped for simplicity. We have
\begin{equation}
k_{\rm uv} = \frac{\rho -1}{r^2_{\rm max}}.
\end{equation}
For our desktop and ODIS images, the dimension of the images is $320 \times 240$ in pixel. Further, the distortion
experienced by these two cameras is a barrel distortion with $r_d < r$. Focusing on the barrel distortion by
considering $\rho \in [0, \frac{1}{4}, \frac{1}{2}, \frac{3}{4}, 1]$, we have
\begin{equation} \label{eqn: k range one coeff}
k_{\rm uv} = \left\{\matrix{%-2.5\times 10^{-5}, & {\rm when} \; \rho = 0, \cr -1.875\times 10^{-5}, & {\rm when} \;
%\rho = \frac{1}{4},\cr -1.25\times 10^{-5}, & {\rm when} \; \rho = \frac{1}{2},\cr
-6.25\times 10^{-6}, & {\rm when} \;
\rho = \frac{3}{4}, \cr 0, & {\rm when} \; \rho = 1.} \right.
\end{equation}
The initial searching range for the distortion coefficients when using the radial distortion function (\ref{eqn: one
coeff}) is chosen to be within $[-4.5\times 10^{-6} \sim 3.5 \times 10^{-6}]$ with a step size $10^{-6}$. Similarly,
for the radial distortion function (\ref{eqn: simplified geometric 6}) with $k_{\rm uv1} = k_{\rm uv2}$, we have
\begin{equation}
k_{\rm uv} = \frac{\frac{1}{\rho} -1}{r^2_{\rm max}},
\end{equation}
and
\begin{equation} \label{eqn: k range 6}
k_{\rm uv} = \left\{\matrix{%\infty, & {\rm when} \; \rho = 0,\cr 7.5 \times 10^{-5}, & {\rm when} \; \rho =
%\frac{1}{4},\cr 2.5\times 10^{-5}, & {\rm when} \; \rho = \frac{1}{2},\cr
8.3\times 10^{-6}, & {\rm when} \; \rho =
\frac{3}{4}, \cr 0, & {\rm when} \; \rho = 1.} \right.
\end{equation}
The initial searching range when using function (\ref{eqn: simplified geometric 6}) is $[0 \sim 9 \times 10^{-6}]$.

\begin{figure}[htb]
\centering
\includegraphics[width=0.4\textwidth]{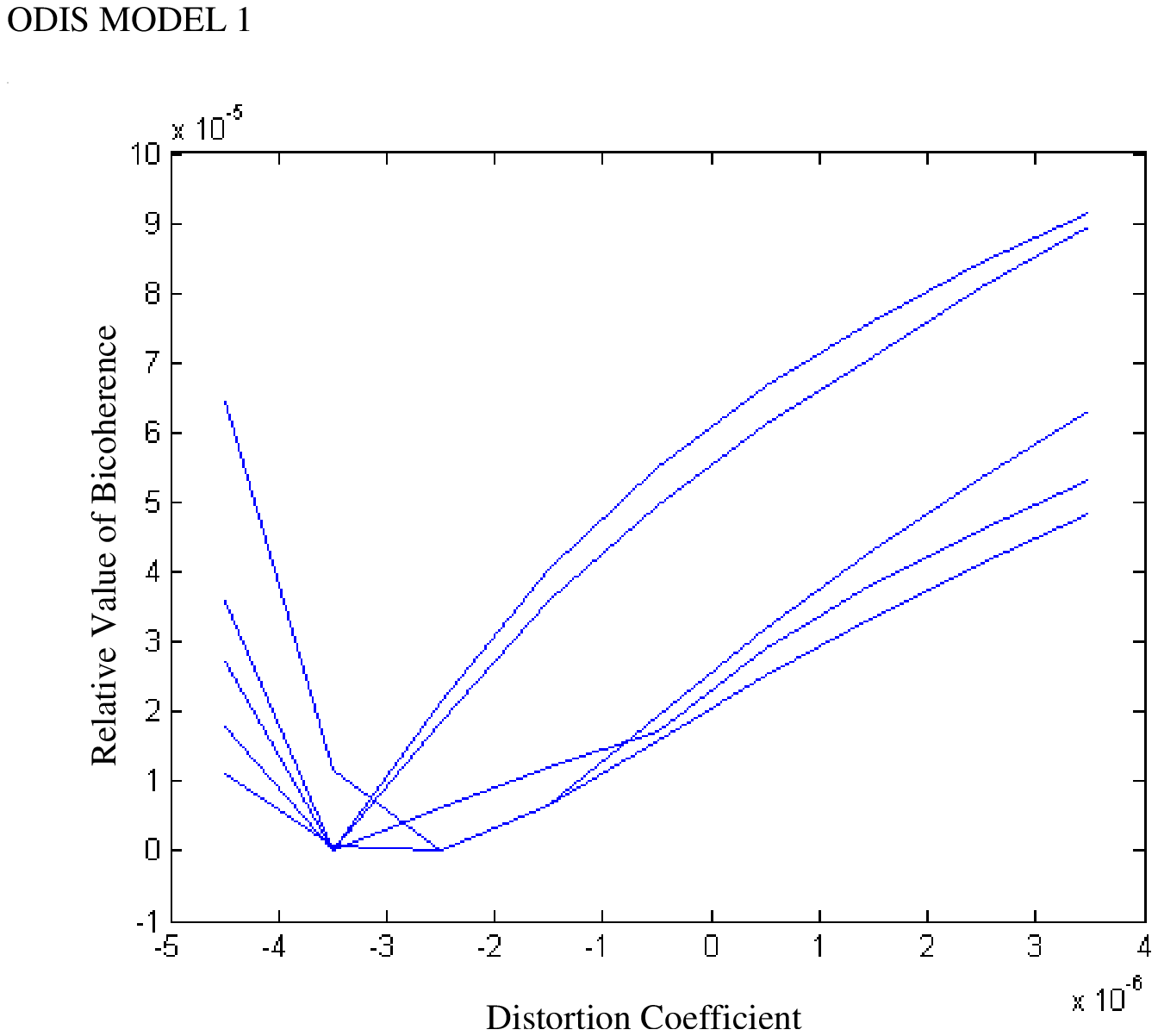}
\caption {Relative $J$ values of the ODIS images using function (\ref{eqn: one coeff}).} \label{fig: Js ODIS model 1}
\end{figure}

% J curves:
Relative values of $J$ as defined in (\ref{eqn: max bicoherence measure}) of the five ODIS images using the distortion
functions (\ref{eqn: one coeff}) and (\ref{eqn: simplified geometric 6}) with $k_{uv1} = k_{uv2}$ are shown in
Figs.~\ref{fig: Js ODIS model 1} and \ref{fig: Js ODIS model 2}. The relative $J$ values equal to their corresponding
values minus the minimum value in this group.

\begin{figure}[htb]
\centering
\includegraphics[width=0.4\textwidth]{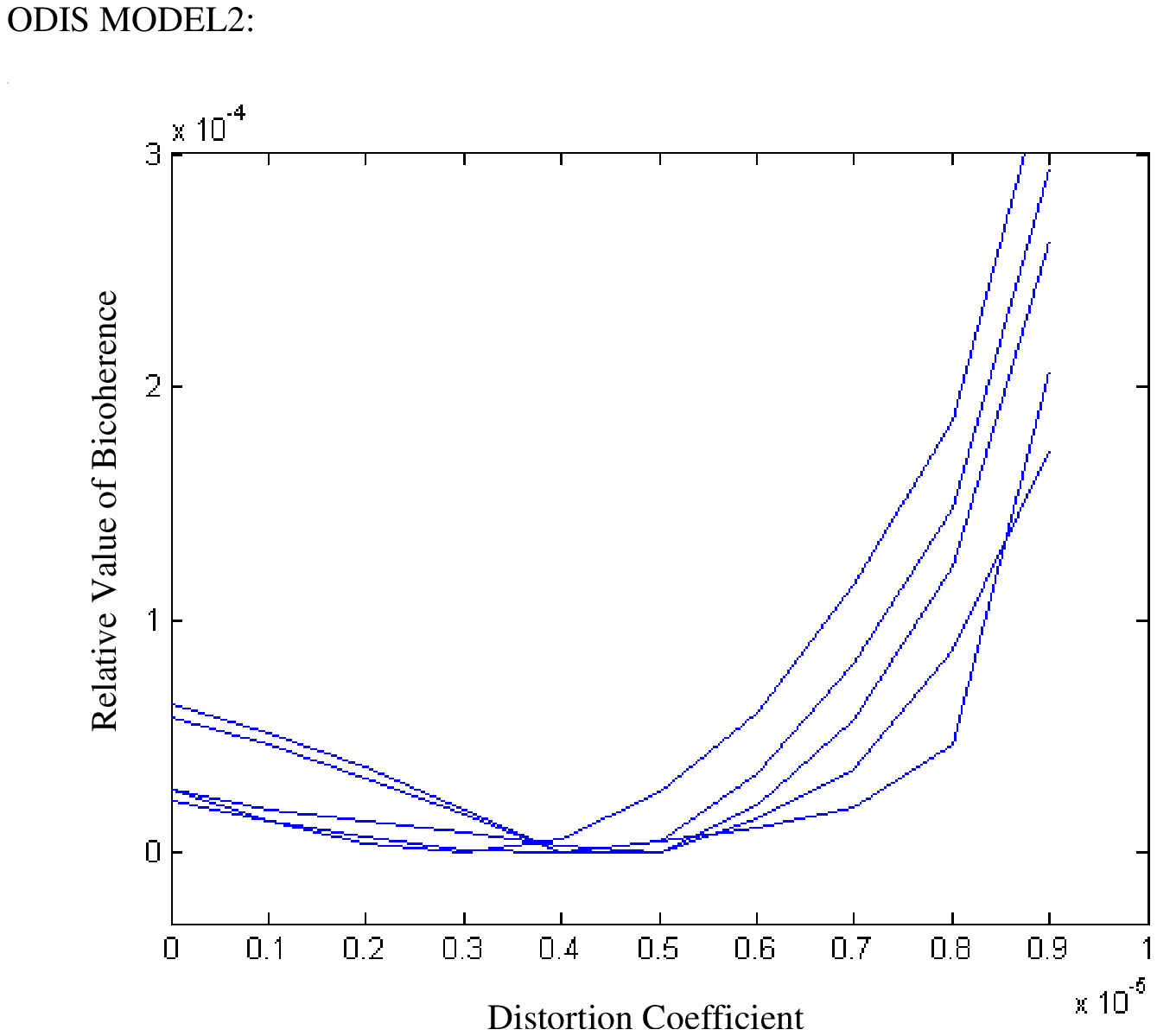}
\caption {Relative $J$ values of the ODIS images using function (\ref{eqn: simplified geometric 6}).} \label{fig: Js
ODIS model 2}
\end{figure}

% ===========================================================================
\subsection{Blind Detection and Compensation of Lens Geometric Nonlinearity}
\label{sec: blind geometric}

Besides the radial distortion modelling, a simplified geometric distortion modelling method has been developed in
\cite{LiliSub2IT03geometric}. A straightforward question asked is whether the higher-order correlation detection in the
frequency domain can help for the detection of geometric distortions, instead of just the radial one. The above problem
is pursued in the next, where we would like to first describe our conclusion. That is, the blind lens removal technique
helps to detect the possible geometric distortion of a camera. By allowing the two distortion coefficients along the
two image axes searched separately in two regions, the blindly calibrated lens distortion coefficients manifest
noticeable difference for cameras that have been reported to be more accurately modelled by a geometric distortion
modelling method.

Again using the two calibrated desktop and the ODIS cameras, the blindly calibrated distortion coefficients $k_{\rm
uv1}$ and $k_{\rm uv2}$ using the rational distortion function (\ref{eqn: simplified geometric 6}) are shown in
Table~\ref{table: blind geometric 6}, where the difference between the two distortion coefficients is significant for
the ODIS camera, which has been studied in \cite{LiliSub2IT03geometric} to be better modelled by a geometric distortion
model than a radial one. It is also observed that this difference between distortion coefficients along the two image
axes is exaggerated using the blind removal technique.

\begin{table}[htb]
\centering \caption{Blind Detection of Geometric Distortion} \label{table: blind geometric 6}
\renewcommand{\arraystretch}{1.3}
\vspace{-2mm} {{\begin {tabular}{|c|c|c|}\hline

               &{\bf Desktop} ($10^{-6}$) & {\bf ODIS} ($10^{-6}$) \\[-1ex]
               & $[k_{\rm uv1}, k_{\rm uv2}]$ & $[k_{\rm uv1}, k_{\rm uv2}]$ \\\hline

               & $[6.16, 5.28]$     & $[2.64, 6.16]$ \\
               & $[2.64, 4.40]$     & $[4.40, 6.16]$ \\
\raisebox{2.0ex}[0pt]{\bf Distortion}   & $[5.28, 6.16]$ & $[2.64, 4.40]$ \\
\raisebox{2.0ex}[0pt]{\bf Coefficients} & $[2.64, 3.52]$ & $[5.28, 6.16]$ \\
               & $[5.28, 3.52]$     & $[3.52, 6.16]$ \\\hline

\bf Mean       & $[4.40, 4.57]$     & $[3.70, 5.81]$ \\\hline
\end {tabular}}}
\end{table}

In our implementation, geometric undistortion is implemented using equation (\ref{eqn: simplified geometric 6}) for
each slice passing through the image center, which is chosen to be 12 degree apart for each image\footnote{Four images
are reported to be enough to output unbiased distortion coefficients \cite{Hany01blindremoval}. In this paper, five
images of each camera are used.}. After $(u,v)$ are derived from $(u_d, v_d)$ and $(k_{\rm uv1}, k_{\rm uv2})$, where
the $(k_{\rm uv1}, k_{\rm uv2})$ are determined through the searching procedures, $r_{\rm uv}$ is calculate from
$(u,v)$ and the image center (which is assumed to be the center of distortion in the context of blind lens distortion
removal technique). Nonlinearity detection using bicoherence is performed on this one-dimensional signal $r_{\rm uv}$,
which is basically the same procedures used in the blind removal of radial distortion.

Due to the available knowledge of the distortion coefficients for the two sets of images using the radial distortion
modeling with function (\ref{eqn: simplified geometric 6}) for $k_{uv1} = k_{uv2}$, when performing the detection for
possible geometric distortions, the searching ranges are chosen to be around the distortion coefficients already
determined in the radial case.

%% ===========================================================================
%\subsection{Blind Lens Compensation Using Web Images}
%
%The blind lens removal technique is used for the radial distortion compensation of the following two web images, where
%the original and the compensated images are plotted in Figs.~\ref{fig: song1} and \ref{fig: song2}\footnote{The
%original images are resized with a scala $1/6$ to reduce the search time.}, respectively. Since the original images in
%Figs.~\ref{fig: song1} and \ref{fig: song2} are captured from the same digital camera, the radial distortion
%coefficient can be searched using one image, instead of multiple images as the case for an analog CCD camera.
%
%\begin{figure}[htb]
%\centering
%\includegraphics[width=0.5\textwidth]{song1}
%\caption {Example of blind radial distortion compensation of a web image (left: original; right: compensated).}
%\label{fig: song1}
%\end{figure}
%
%
%\begin{figure}[htb]
%\centering
%\includegraphics[width=0.5\textwidth]{song2}
%\caption {Example of blind radial distortion compensation of a web image (left: original; right: compensated).}
%\label{fig: song2}
%\end{figure}
%

%%%%%%%%%%%%%%%%%%%%%%%%%%%%%%%%%%%%%%%%%%%%%%%%%%%%%%%%%%%%%%%%%%%%%%%%%%%%%%%%%%%%%%%%%%%%%%%%%%%%%%%%%%%%
\section{Concluding Remarks and Discussions}
\label{sec: conclusions}

Higher-order correlation based technique is a promising method to detect lens nonlinearities in the absence of the
camera. Besides the commonly used single-coefficient polynomial radial distortion model, blind detection of the
geometric distortion is addressed in this paper. Using the quantitative measurement criterion defined in the frequency
domain, difference between the two sets of distortion coefficients along the two image axes can be used to detect
cameras that are better modelled by a geometric distortion modelling method qualitatively.

%\section{Acknowledgement}
%The authors would like to thank Zhen Song and his father for providing the original images in Figs.~\ref{fig: song1}
%and \ref{fig: song2}.

%%%%%%%%%%%%%%%%%%%%%%%%%%%%%%%%%%%%
\bibliography{d:/bibs/calibration}

\begin{thebibliography}{10}

\bibitem{OlivierF01Straight}
Frederic Devernay and Olivier Faugeras,
\newblock ``Straight lines have to be straight,''
\newblock {\em Machine Vision and Applications}, vol. 13, no. 1, pp. 14--24,
  2001.

\bibitem{tsai87AVersatile}
Roger~Y. Tsai,
\newblock ``A versatile camera calibration technique for high-accuracy 3{D}
  machine vision metrology using off-the-shelf {TV} cameras and lenses,''
\newblock {\em IEEE Journal of Robotics and Automation}, vol. 3, no. 4, pp.
  323--344, Aug. 1987.

\bibitem{Photogrammetry80}
Chester~C Slama, Ed.,
\newblock {\em Manual of Photogrammetry},
\newblock American Society of Photogrammetry, fourth edition, 1980.

\bibitem{zhang99calibrationinpaper}
Zhengyou Zhang,
\newblock ``Flexible camera calibration by viewing a plane from unknown
  orientation,''
\newblock {\em IEEE Int. Conf. on Computer Vision}, pp. 666--673, Sept. 1999.

\bibitem{Heikkil97fourstepcameracalibration}
J.~Heikkil and O.~Silvn,
\newblock ``A four-step camera calibration procedure with implicit image
  correction,''
\newblock in {\em IEEE Computer Society Conference on Computer Vision and
  Pattern Recognition}, San Juan, Puerto Rico, 1997, pp. 1106--1112.

\bibitem{Janne96Calibration}
Janne Heikkila and Olli Silven,
\newblock ``Calibration procedure for short focal length off-the-shelf {CCD}
  cameras,''
\newblock in {\em Proceedings of 13th International Conference on Pattern
  Recognition}, Vienna, Austria, 1996, pp. 166--170.

\bibitem{Undistortionchapter}
Charles Lee,
\newblock {\em Radial Undistortion and Calibration on An Image Array},
\newblock Ph.D. thesis, MIT, 2000.

\bibitem{Juyang92distortionmodel}
Juyang Weng, Paul Cohen, and Marc Herniou,
\newblock ``Camera calibration with distortion models and accuracy
  evaluation,''
\newblock {\em IEEE Trans. on Pattern Analysis and Machine Intelligence}, vol.
  14, no. 10, pp. 965--980, Oct. 1992.

\bibitem{Toru02Unified}
Toru Tamaki, Tsuyoshi Yamamura, and Noboru Ohnishi,
\newblock ``Unified approach to image distortion,''
\newblock in {\em International Conference on Pattern Recognition}, Aug. 2002,
  pp. 584--587.

\bibitem{Hany01blindremoval}
Hany. Farid and Alin~C. Popescu,
\newblock ``Blind removal of lens distortion,''
\newblock {\em Journal of the Optical Society of America A, Optics, Image
  Science, and Vision}, vol. 18, no. 9, pp. 2072--2078, Sept. 2001.

\bibitem{Mendel91HOS}
Jerry~M. Mendel,
\newblock ``Tutorial on higher-order statistics (spectra) in signal processing
  and system theory: Theoretical results and some applications,''
\newblock {\em Proceedings of the IEEE}, vol. 79, no. 3, pp. 278--305, Mar.
  1991.

\bibitem{LiliSub2IT03geometric}
Lili Ma, YangQuan Chen, and Kevin~L. Moore,
\newblock ``A family of simplified geometric distortion models for camera
  calibration,'' {\tt http://} {\tt arxiv.org/} {\tt abs/cs.CV/0308003}, 2003.

\bibitem{Lilicalreport02}
Lili Ma,
\newblock ``Camera calibration: a {USU} implementation,'' CSOIS Technical
  Report, ECE Department, Utah State University, {\tt
  http://arXiv.org/abs/cs.CV/0307072}, May, 2002.

\bibitem{odiscamera}
``Cm3000-l29 color board camera ({ODIS} camera) specification sheet,'' {\tt
  http} {\tt://www.} {\tt video-} {\tt surveillance} {\tt-hidden} {\tt-spy}
  {\tt-cameras}{\tt .com/} {\tt cm3000l29.htm}.

\end{thebibliography}
%%%%%%%%%%%%%%%%%%%%%%%%%%%%%%%%%%%%
\end{document}